\documentclass[final]{cvpr}

\usepackage{times}
\usepackage{epsfig}
\usepackage{graphicx}
\usepackage{amsmath}
\usepackage{amssymb}
\usepackage{bm}

\usepackage[labelsep=period]{caption}
\usepackage{float}
\DeclareMathOperator*{\argmax}{arg\,max}

\usepackage[normalem]{ulem}
\usepackage{booktabs, multirow} 
\usepackage{soul}
\usepackage[table]{xcolor} 
\usepackage{changepage,threeparttable} 
\usepackage{arydshln}

\usepackage[symbol]{footmisc}
\renewcommand{\thefootnote}{\fnsymbol{footnote}}

\usepackage[pagebackref=true,breaklinks=true,colorlinks,bookmarks=false]{hyperref}

\def\cvprPaperID{5224} 

\def\cvprPaperID{5224} 
\def\confYear{CVPR 2021}

\title{Pixel-wise Anomaly Detection in Complex Driving Scenes}

\author{Giancarlo Di Biase\footnote[1]{Equal Contribution}\\
ETH Zurich\\
{\tt\small giandbt@gmail.com}
\and
Hermann Blum\footnote[1]{}\\
ETH Zurich\\
{\tt\small hermann.blum@ethz.ch}

\and
Roland Siegwart\\
ETH Zurich\\
{\tt\small rsiegwart@ethz.ch}

\and
Cesar Cadena\\
ETH Zurich\\
{\tt\small cesarc@ethz.ch}
}

\begin{document}

\twocolumn[{%
\renewcommand\twocolumn[1][]{#1}%
\maketitle
\begin{center}
    \vspace{-5mm}
   \centering
    \includegraphics[width=1.0\textwidth]{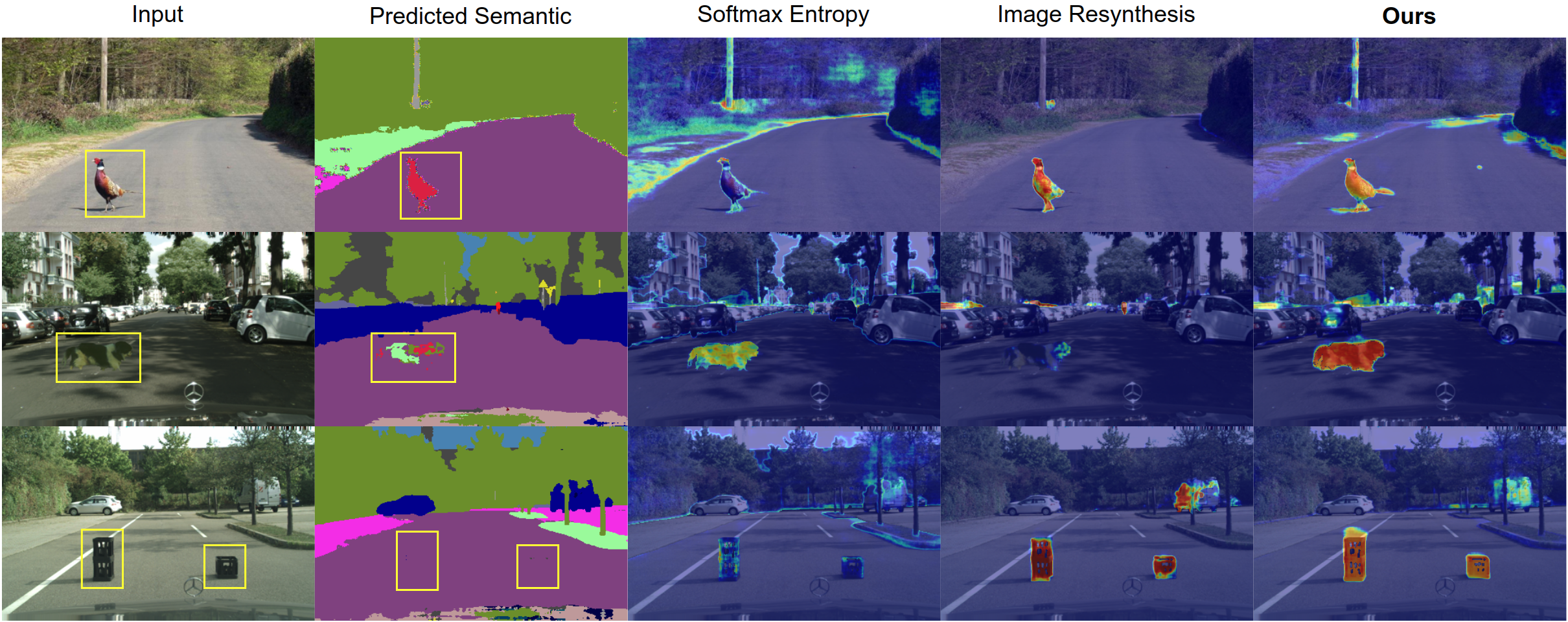}
    \captionof{figure}{\textbf{Anomaly scenarios overview.} There are three possible outcomes when a segmentation network encounters an anomalous instance. First, anomaly instances are properly segmented and classified as one of the training classes (i.e bird is confused as a person) (top). Second, anomaly instances are over-segmented with multiple classes (i.e dog is detected as a combination of person, vegetation, and terrain classes) (middle). And third, anomaly instances are blended with the background, not detected (i.e boxes blend with the street segmentation) (bottom). Our proposed method produces robust predictions for all scenarios, while previous approaches fail to handle at least one of them.}
    \label{fig:anomaly_scenarios}
    \vspace{-1mm}
\end{center}%
}]

\begin{abstract}
   The inability of state-of-the-art semantic segmentation methods to detect anomaly instances hinders them from being deployed in safety-critical and complex applications, such as autonomous driving.
   Recent approaches have focused on either leveraging segmentation uncertainty to identify anomalous areas or re-synthesizing the image from the semantic label map to find dissimilarities with the input image.
   In this work, we demonstrate that these two methodologies contain complementary information and can be combined to produce robust predictions for anomaly segmentation. 
   We present a pixel-wise anomaly detection framework that uses uncertainty maps to improve over existing re-synthesis methods in finding dissimilarities between the input and generated images. 
   Our approach works as a general framework around already trained segmentation networks, which ensures anomaly detection without compromising segmentation accuracy, while significantly outperforming all similar methods.  
  Top-2 performance across a range of different anomaly datasets shows the robustness of our approach to handling different anomaly instances.
   
   \vspace{-3mm}
\end{abstract}
\let\thefootnote\relax\footnotetext{\hspace{-6mm}* Equal Contribution\\
This work was partially supported by the Hilti Group and the National Center of Competence in Research (NCCR) Robotics through the SwissNational Science Foundation.%
}

\section{Introduction}



Recent advances in deep learning have shown significant improvements in the field of computer vision. Neural networks have become the de-facto methodology for classification, object detection, and semantic segmentation due to their high accuracy in comparison to previous methods \cite{best_classification, best_objectdetection, best_segmentation}. However, while the predictions of these networks are highly accurate, they usually fail when encountering anomalous inputs (i.e. instances outside the training distribution of the network).

With this work, we focus on the inability of existing semantic segmentation models to localize anomaly instances and how this limitation hinders them from being deployed in safety-critical, in-the-wild scenarios. 
Consider the case of a self-driving vehicle that uses a semantic segmentation model. If the agent encounters an anomalous object (i.e. a wooden box in the middle of the street), the model could wrongly classify this object as part of the road and lead the vehicle to crash.

To detect such anomalies in the input, we build our approach  upon two established groups of methods.
The first group uses uncertainty estimation to detect anomalies. Their intuition follows that a low-confidence prediction is likely an anomaly. However, uncertainty estimation methods themselves are still noisy and inaccurate.  Previous works \cite{epfl,fishyscapes} have shown that these models fail to detect many unexpected objects. Example failure cases are shown in Figure~\ref{fig:anomaly_scenarios} (top and bottom) where the anomaly object is either detected but miss-classified or non-detected and blended with the background. In both cases, the segmentation network is overconfident about its prediction and, thus, the estimated uncertainty (softmax entropy) is low.   

The second group focuses on re-synthesizing the input image from the predicted semantic map and then comparing the two images (input and generated) to find the anomaly. These models have shown promising results when dealing with segmentation overconfidence but fail when the segmentation outputs a noisy prediction for the unknown object, as shown in Figure \ref{fig:anomaly_scenarios} (middle). This failure is explained by the inability of the synthesis model to reconstruct noisy patches of the semantic map, which complicates finding the differences between input and synthesized images.

In this paper, we propose a novel pixel-level anomaly framework that combines uncertainty and re-synthesis approaches in order to produce robust predictions for the different anomaly scenarios. Our experiments show that uncertainty and re-synthesis approaches are complementary to each other, and together they cover the different outcomes when a segmentation network encounters an anomaly.

Our framework builds upon previous re-synthesis methods \cite{epfl, master_thesis_old, synthesize_compare} of reformulating the problem of segmenting unknown classes as one of identifying differences between the input image and the re-synthesised image from a predicted semantic map. We improve over those frameworks by integrating different uncertainty measures, such as softmax entropy \cite{entropy, bayes_ensemble}, softmax difference \cite{meta}, and perceptual differences \cite{perceptual_diff, perceptual_diff_2} to assist the dissimilarity network in differentiating the input and generated images.
The proposed framework successfully generalizes to all anomalies scenarios, as shown in Figure \ref{fig:anomaly_scenarios}, with minimal additional computation effort and without the need to jeopardize the segmentation network accuracy (no re-training necessary), which is one common flaw of other anomaly detectors \cite{anomaly_score, prior_entropy, bayes_deep}. Besides maintaining state-of-the-art performance in segmentation, eliminating the need for re-training also reduces the complexity of adding an anomaly detector to future segmentation networks, as training these networks is non-trivial. 

We evaluate our framework\footnote{Available at \url{https://github.com/giandbt/SynBoost}.} in public benchmarks for anomaly detection, where we compare to methods similar to ours that not compromise segmentation accuracy, as well as those requiring full retraining. 
We also demonstrate that our framework is able to generalize to different segmentation and synthesis networks, even when these models have lower performance. We replace the segmentation and synthesis models with lighter architectures to prioritize speed in time-critical scenarios like autonomous driving. 

 
In summary, our contributions are the following:
\begin{itemize}
    \item[--] We present a novel pixel-wise anomaly detection framework that leverages the best features of existing uncertainty and re-synthesis methodologies.
    \item[--] Our approach is robust to the different anomaly scenarios, achieving state-of-the-art performance on the Fishyscapes benchmark while maintaining state-of-the-art segmentation accuracy.  
    \item[--] Our proposed framework is able to generalize to different segmentation and synthesis networks, serving as a wrapper methodology to existing segmentation pipelines.
\end{itemize}




\section{Related Work}





The task of localizing anomalous instances in semantic segmentation has been studied under out-of-distribution (OoD) detection and anomaly segmentation. In this section, we review the methods which could be used for pixel-wise anomaly detection, and exclude approaches that could \textit{only} be applied for image-level OoD classification. 

\subsection{Anomaly Segmentation via Uncertainty Estimation}
Methods that estimate the uncertainty of a model for a given input may estimate high uncertainty for inputs that are not anomalies, e.g. due to high input noises. Regardless of this and other differences, anomaly detection is a common benchmark method for uncertainty estimation, based on the assumption that anomalous inputs should come with higher uncertainty than any training data.

Early methods measure uncertainty from the predicted softmax distributions and classify the samples as OoD by using simple statistics \cite{MSP, calibrated_confidence, odin}. While these approaches are good baselines for image-level OoD classification, they usually fail in anomaly segmentation. Specifically, the estimated (aleatoric) uncertainty is often high at object boundaries, where no single label can be assigned with certainty, and not at anomalous instances as desired. \cite{meta} mitigated this shortcoming by aggregating different dispersion measurements (e.g., entropy, and difference in softmax probability) and then predicting areas of potential high error in the segmentation. Then, \cite{meta_ood} demonstrated that these high error areas can be used to localize anomalies by visual feature differences. 

Alternative approaches use Bayesian NNs with MC dropout to estimate pixel uncertainty \cite{bayes_uncertainty, bayes_deep, bayes_ensemble}. 
These methods differentiate between aleatoric (noise inherent in the observations) and epistemic uncertainties (uncertainty in the model),
therefore mitigating the problem of object boundaries, but still fail to detect anomalies on a pixel level accurately. As shown in \cite{epfl}, they yield many false positive predictions, as well as miss-matches between anomaly instances and uncertain areas. 

\subsection{Anomaly Segmentation via Outlier Exposure}
Anomaly segmentation can also be accomplished by training a network to differentiate inliers against unseen samples by using an auxiliary dataset of outliers \cite{outlier_exposure}. \cite{outlier_train} was one of the first approaches to use outlier exposure for dense predictions using ImageNet \cite{imagenet} as the OoD dataset. Then, \cite{anomaly_score} build upon this methodology by modifying the segmentation network to predict the semantic map as well as the outlier map. Note that this requires re-training the segmentation network as a multi-task model, which has lead to drop in performance \cite{multitask}. 
The biggest shortcoming of these approaches is that they train from OoD samples, which could compromise their ability to generalize to all possible anomalies. 

\subsection{Anomaly Segmentation via Image Re-synthesis}
Promising recent methods follow the approach of reconstructing the input image using generative models. The intuition behind this methodology is that the generated image will yield appearance differences with respect to the input image where anomalies are present, as the model cannot handle these instances. Early work on this sub-field used autoencoders to re-synthesize the original image \cite{ae_1, ae_2}. However, these methods mostly generated a lower-quality version of the input image \cite{epfl}. 
More recent methods \cite{synthesize_compare, master_thesis_old, epfl} re-synthesize the input image from the predicted semantic map using a generative adversarial network. The photo-realistic image is then compared to the original image by a discrepancy or comparison module to localize the anomaly instances. 

These approaches benefit from not needing to re-train the segmentation network as they work as a wrapper method. Additionally, they do not require OoD samples which helps them to generalize to never-seen anomalies instances. 
However, the performance of these methods are limited by the ability of the discrepancy module to differentiate between features in the input and generated images, which could be challenging for complex driving scenes. 
With our work, we demonstrate that feeding uncertainty information of the scene to the discrepancy network significantly improves the ability of the module to detect anomalies. 





\section{Methodology}
\label{methodolgy}



\begin{figure*}[!th]
\begin{center}
   \includegraphics[width=0.95\linewidth,height=6.3cm]{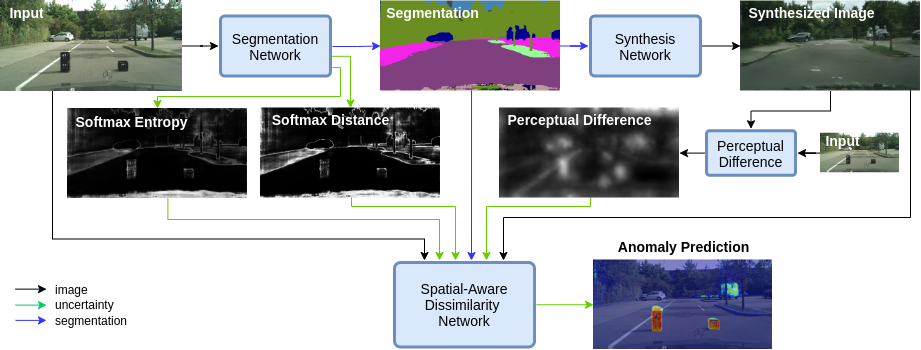}
   \vspace{-5mm}
\end{center}
   \caption{\textbf{Anomaly Segmentation Framework}. We first pass the input image through a segmentation network, which will output a semantic map and two uncertainty maps. The predicted semantic map is then processed by the synthesis network to generate a photo-realistic image. Perceptual difference is then calculated by comparing features between the input and generated images. Lastly, all the predicted images and the input are sent to the spatial-aware dissimilarity module to produce the anomaly prediction.}
\label{fig:framework}
\vspace{-4mm}
\end{figure*}

We propose a detection framework for segmenting anomalous instances. Our framework is inspired by recent re-synthesis approaches \cite{epfl,synthesize_compare}, while extending them to include the benefits of uncertainty estimation methods \cite{meta_ood,meta}. 
We first introduce our framework with its respective modules (Sec. \ref{met:framework}). Next, we describe how to train the modules to better handle the different anomaly scenarios (Sec. \ref{met:training}). Finally, we combine our framework's output with the calculated uncertainty maps to have a final ensemble method that reduces the false positives and overconfidence in the predictions (Sec. \ref{met:ensemble}).


\subsection{Pixel-wise Anomaly Detection Framework}
\label{met:framework}

Our proposed framework follows the same base structure as \cite{epfl} and \cite{master_thesis_old}, where the input image gets segmented, a reconstruction is synthesized from the segmentation map, and a dissimilarity module detects anomalies by comparing input and synthesized image. However, we extended each component to predict and/or use uncertainty measurements to improve the final anomaly prediction. Figure \ref{fig:framework} shows a high-level summary of our framework. 

\subsubsection{Segmentation Module}
The segmentation module takes the input image and feeds it into a segmentation network, such as \cite{DeepLabV3+Label} or \cite{PSPNet}, in order to obtain a semantic map. In addition to the semantic map, we also compute two dispersion measures to quantify the uncertainty in the semantic map prediction. These two dispersion measurements are the \textit{softmax entropy} $H$ \cite{entropy, bayes_ensemble} and the \textit{softmax distance} $D$ (i.e. the difference between the two largest softmax values), which have shown to be beneficial in understanding errors within the segmentation \cite{meta}. For each pixel \textit{x}, these two measurements are calculated as follows:

\begin{equation}
 H_x = - \sum\limits_{c \in \textrm{classes}} {p(c)\: log_2\:p(c)}
 \label{eq:1}
\end{equation}

\begin{equation}
 D_x = 1 - \max_{c \in \textrm{classes}}p(c) + \max_{c'\in \textrm{classes} \setminus{(\argmax_c p(c))} }p(c')
\end{equation}

where $p(c)$ is the softmax probability  for class $c$. We normalize both quantities to $[0,1]$.

\subsubsection{Synthesis Module}
The synthesis module generates a realistic image out of the given semantic map with pixel-to-pixel correspondence. It is trained as a conditional generative adversarial network (cGAN)~\cite{Pix2PixHd, ccfpse} to fit the generative distribution to the distribution of input images from the semantic model.

While the synthesis module is trained to produce photorealistic images and is well able to produce realistic cars, buildings, or pedestrians, the semantic map misses essential information like color or appearance to allow for a direct per-pixel value comparison. We therefore calculate the \textit{perceptual difference} $V$ between the original and synthesized image. This novel feature map is inspired from the perceptual loss presented in \cite{perceptual_diff} and \cite{perceptual_diff_2}, which is commonly used in cGANs methods. 
The idea is to find the pixels that have the most different feature representations using ImageNet \cite{imagenet} pre-trained VGG as feature extractor \cite{vgg}. The difference in these representations allow us to compare objects based on their image content and spatial structure, as opposed to low level features such as color and texture. If the anomaly object is not detected or wrongly classified, the synthesized image would be generated with the wrong feature representation, and thus the perceptual difference should detect these discrepancies with the input image. 

For every pixel $x$ of the input image and corresponding pixel $r$ from the synthesized image, the perceptual difference is calculated as follows:

\begin{equation}
V(x,r) = \sum_{i=1}^{N} \dfrac{1}{M_i} \left\Vert F^{(i)}(x) - F^{(i)}(r)\right\Vert_1 
\end{equation}

where $F^{(i)}$ denotes the $i$-th layer with $M_i$ elements of the VGG network and $N$ layers. For consistency, these dispersion measure is also normalized between $[0,1]$. 

\begin{figure*}[!th]
\begin{center}
   \includegraphics[width=1.0\linewidth, height=7.5cm]{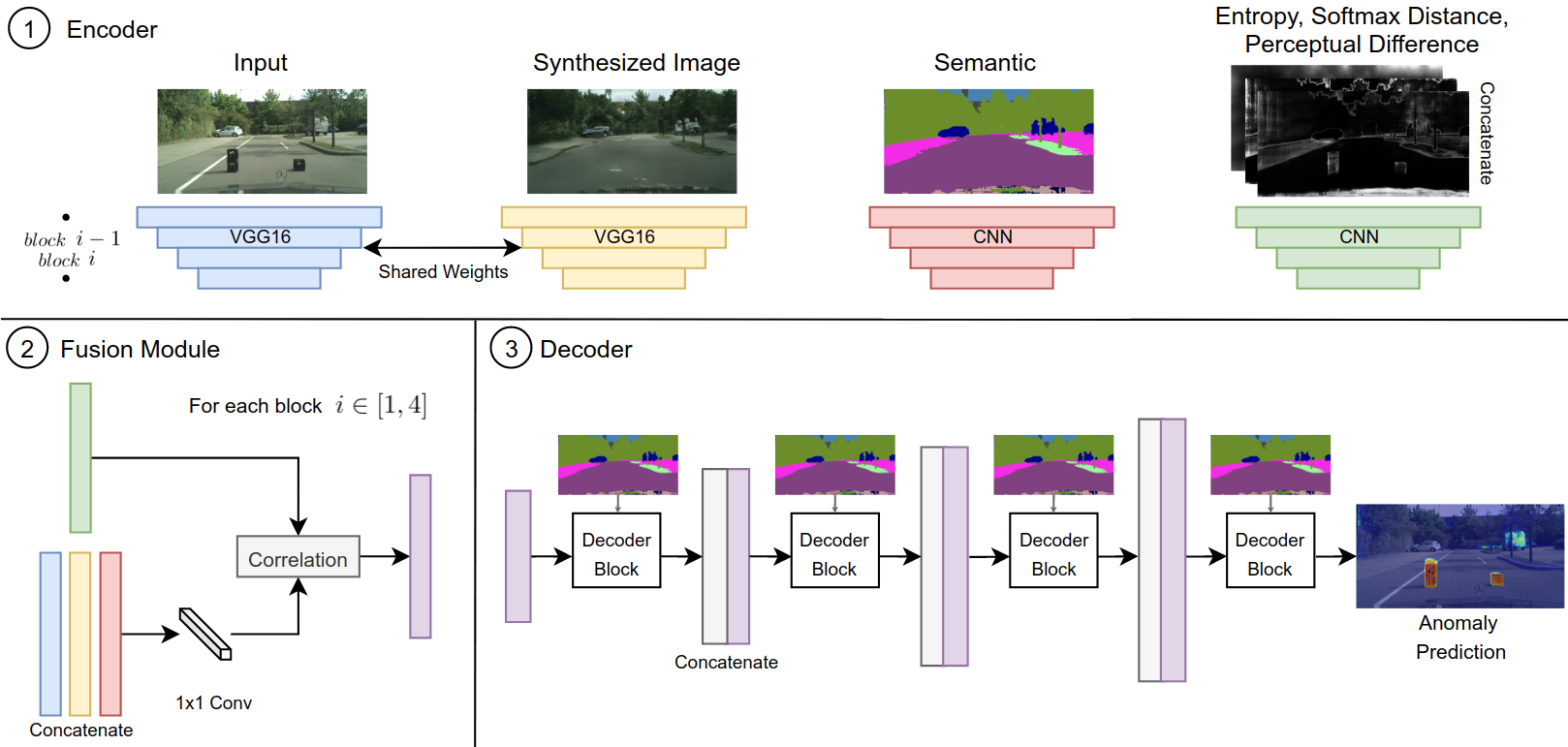}
   \vspace{-9mm}
\end{center}
   \caption{\textbf{Dissimilarity Module Architecture}. Given the input, synthesized, semantic, and uncertainty images, we extract high to low-level features for each image with a CNN. For each level, we then concatenate the input (blue), synthesized (yellow), and semantic (red) features and fuse them with a 1x1 convolution. The resulting map is used to calculate a correlation with the features from the uncertainty maps (green). The output of the fusion module (purple) is then fed to a decoder to produce the predicted anomaly segmentation. Note the semantic map is used in the decoder block to ensure a spatial aware prediction by using a SPADE normalization \cite{SPADE}.}
\label{fig:dissimilarity}
\vspace{-5mm}
\end{figure*}

\subsubsection{Dissimilarity Module}
The dissimilarity module takes as input the original image, generated image, and semantic map, as well as the uncertainty maps (softmax entropy, softmax distance, perceptual difference) calculated in previous steps. Then, the network combines these features to predict the anomaly segmentation map. The dissimilarity module is divided into three components (encoder, fusion module, and decoder) as seen in Figure \ref{fig:dissimilarity}. Implementation details can be found in Appendix~\ref{app:dissimilarity}.

\textbf{Encoder}. Each image is passed through an encoder to extract features. Similarly to \cite{epfl}, we use a pre-trained VGG \cite{vgg} network for both the original and re-synthesized images as well as a simple CNN to process the semantic map. We also added another simple CNN to encode the uncertainty maps, which are all concatenated. 

\textbf{Fusion Module}. At each level of the feature pyramid, we concatenate the input, synthesis, and semantic feature maps and pass them through a $1\times 1$ convolution. With this first step, we are training the network to differentiate between the original and generated images, as it is common for re-synthesis methods. Additionally, we use the resulting feature map and perform a point-wise correlation with the uncertainty feature map. This step guides the network to pay attention to high-uncertain areas in the feature map. 

\textbf{Decoder}. We then decode each feature map and concatenate it with the corresponding higher level in the pyramid until we get our anomaly segmentation prediction. Note that the decoder block uses the semantic map as an input since it uses a spatial-aware normalization as presented in \cite{SPADE}. This normalization was used to ensure semantic information is not wash-away during the decoding process.    

\subsection{Training Procedure}
\label{met:training}

Anomaly instances by definition include any object which does not belong to the training classes. As such, it is crucial to ensure the proposed methodology is robust to detect any anomaly and does not overfit to specific objects from an OoD dataset. 
Additionally, the training needs to be general to cover all three anomaly scenarios of Figure \ref{fig:anomaly_scenarios}.

The segmentation and synthesis module are simply trained on the inlier segmentation dataset. This is therefore free of any assumptions on anomalies, and also ensures that the segmentation module is solely trained on the segmentation task without balancing in any other factors.

To train the dissimilarity module, 
\cite{epfl} solved the problem of not needing OoD dataset by generating synthetic data to simulate segmentation maps in the presence of anomalies. This method replaced the class of randomly-chosen object instances in the ground truth semantic map for an alternative random class. Then, these altered semantic maps are synthesized (with the already trained module), creating visual differences between real and generated images. These differences are used by the dissimilarity network to train and detect discrepancies.  
Even though this approach does not require seen OoD objects during training, it falls short to train a fully robust pixel-wise anomaly detection. 
First, this training data generator only covers one anomaly scenario (Figure \ref{fig:anomaly_scenarios} - top) but lacks examples for the other two. 
Second, this approach trains the dissimilarity network using ground truth semantic maps, but during inference, it uses predicted maps. This change in distribution can negatively affect the performance of the network to detect anomalies. 
Lastly, specific to the methodology presented in Sec. \ref{met:framework}, the altered instances would not correlate to uncertain areas. As such, these training examples cannot leverage information from uncertainty maps. 

We expanded the training data generator by adding a second source of training examples.
Specifically, we label objects within the void class in ground truth semantic maps as anomalies. The void class is commonly used in segmentation datasets \cite{Cityscapes} to cover objects that do not belong to any of the training classes, which falls within the definition of an anomaly. 
This additional source of labels helped us to overcome the challenges of the previous data generator. 
First, the predicted semantic map in these void regions will closely match the three anomaly scenarios seen during inference (Figure \ref{fig:anomaly_scenarios}), as opposed to just one. 
Second, the dissimilarity network will train using predicted semantic maps as opposed to ground truth which prevents any domain adaption during inference. 
Lastly, void regions usually match with high uncertainty pixels which guide the dissimilarity network to use uncertainty information.

Note that by adding these void class objects as anomalies, we lose the benefit of not requiring OoD data during training. However, by using both approaches, our proposed methodology is still able to generalize to unseen objects as shown in Sec. \ref{exp:results}.


\subsection{Predictions Ensemble}
\label{met:ensemble}
Until now, we have used the calculated uncertainty maps (i.e. softmax entropy, softmax distance and perceptual difference) as an attention mechanism in the dissimilarity module. 
However, as shown in \cite{meta} and \cite{meta_ood}, these uncertainty maps are already anomaly predictions by themselves. 
Thus, we can exploit the uncertainty estimates to complement the output of the dissimilarity network in order to detect all the anomaly scenarios. 
With this ensemble, we mitigate any possible overconfidence of the dissimilarity network, a common problem with deep learning models as explained by \cite{Calibration}. 

We ensemble these predictions using a weighted average, where the weights are selected by grid search. 
Additionally, we tested learning the ensemble weights in an end-to-end training of the dissimilarity network. These results were not satisfactory as the network still produces overconfident predictions. Further details about end-to-end training can be found under Appendix \ref{app:end-to-end}.
\section{Experiments}
\label{experiments}



\subsection{Experiments Set-up}

\label{exp:set_up}

\textbf{Module Implementations}. The segmentation and synthesis modules use publicly available state-of-the-art networks already trained on Cityscapes \cite{Cityscapes}. Specifically, the segmentation module uses the work presented by \cite{DeepLabV3+Label}, while the synthesis module uses the generator trained by \cite{ccfpse}. A full description of the dissimilarity network implementation can be found in Appendix \ref{app:dissimilarity}. It is important to note that the segmentation module uses the original Cityscapes resolution of 2048 x 1024 as its input, the synthesis module downsamples the resolution by two and finally, the dissimilarity module downsamples the original resolution by four. This downscaling was done due to GPU memory constrains and for faster inference time. 

\textbf{Datasets}. We evaluated the performance of our framework with the Fishyscapes benchmark \cite{fishyscapes}. Fishyscapes is a public benchmark for uncertainty/anomaly estimation in semantic segmentation for urban driving. The benchmark is divided into three sets: FS Lost \& Found (L\&F), FS Static and FS Web. For all datasets, we provide qualitative evaluations on the public validation images, but submitted our method to the benchmark for quantitative results on the private test sets.
FS Lost \& Found is a set of real images captured in~\cite{LostFound} with anomalous objects in front of the vehicle. The test set has 275 images.
FS Static blends anomalous objects from Pascal VOC \cite{pascal-voc} into validation images from Cityscapes. The test set has 1,000 images.
FS Web is a dynamically changing dataset, which overlay objects crawled from the internet using a list of keywords. Methods are only evaluated on data that is crawled after submission to the benchmark, which is why we can only report our method's performance on FS Web Oct. 2020.  

Note that there is another anomaly segmentation benchmark named Street Hazards \cite{streethazards}, which some related works have used to evaluate their performance in pixel-wise anomaly detection \cite{synthesize_compare}. Unfortunately, this benchmark is not compatible with our data generation and training procedure. As explained in Sec. \ref{met:training}, our method requires instance labels, as well as a void class to successfully learn to differentiate the synthesis and original image. The training dataset of Street Hazards  does not provide either. We therefore could not evaluate on the Street Hazards Benchmark.

\textbf{Evaluation Metrics}.
To assess the performance of the framework against existing methods, we use the same metrics presented in the Fishyscapes benchmark for anomaly detection: average precision (AP) and the false positive rate at 95\% true positive rate (FPR95). It is important to note that previous works have used the receiver operating curve (ROC) as their metric for anomaly segmentation (\cite{epfl}, \cite{synthesize_compare}). Nonetheless, ROC is not well-suited for highly imbalance problems, such as anomaly detection, as explained in \cite{roc}.

\begin{table*}[]\centering
\setlength{\tabcolsep}{3.5pt}
\begin{tabular}{clccccccc}\toprule
&\multirow{2}{*}{\textbf{Method}} &\multicolumn{2}{c}{\textbf{FS L\&F}} &\multicolumn{2}{c}{\textbf{FS Static}} &\multicolumn{2}{c}{\textbf{FS Web Oct. 2020}} &\textbf{CS} \\
\textbf{} & &$\uparrow$AP &$\downarrow$FPR95 &$\uparrow$AP &$\downarrow$FPR95   &$\uparrow$AP &$\downarrow$FPR95 &$\uparrow$mIOU \\\toprule
\multirow{4}{*}{\rotatebox{90}{\textit{No Retrain}}} &Softmax Entropy \cite{MSP} &2.93 &44.83 &15.41 &39.75 &16.61 &39.79 &80.30 \\
&Embedding Density \cite{fishyscapes} &4.65 &24.36  &62.14 &\textbf{17.43} &29.16 &38.80 &80.30 \\
&Image Resynthesis++ \cite{epfl} &5.70 &48.05 &29.60 &27.13 &12.46 &51.29 &\textbf{83.50}  \\ 
&\emph{Ours} &\emph{\textbf{43.22}} &\emph{\textbf{15.79}} &\emph{\textbf{72.59}} &\emph{18.75} &\emph{\textbf{61.31}} &\emph{\textbf{18.89}} &\emph{\textbf{83.50}} \\\midrule
\multirow{3}{*}{\rotatebox{90}{\textit{Retrain}}} &Bayesian DeepLab \cite{bayes_deep} &9.81 &38.46 &48.70 &15.50 &35.80 &25.67 &73.80 \\
&Dirichlet DeepLab \cite{prior_entropy} &34.28 &47.43 &31.30 &84.60 &30.02 &76.62 &70.50 \\
&Outlier Head \cite{anomaly_score} &30.92 &22.18 &\textbf{84.02} &\textbf{10.34} &\textbf{63.99} &\textbf{18.79} &77.30 \\
\bottomrule
\end{tabular}
\vspace{-2mm}
\caption{\textbf{Comparison between anomaly segmentation methods}. Our method achieves higher AP and lower FPR95 than previous methods that do not compromise segmentation performance (class mIOU on Cityscapes). It also achieves second-best performance when compared to all existing approaches.}
\label{tab:results}
\vspace{-5mm}
\end{table*}


\textbf{Baselines}.
We compare our approach against all existing methods shown in the Fishyscapes benchmark\footnote{Results for the Fishyscapes benchmark can be found here: \url{https://fishyscapes.com/results}.}.
As of November 2020, the benchmark includes Dirichlet DeepLab \cite{prior_entropy}, Outlier Head \cite{anomaly_score}, Bayesian DeepLab \cite{bayes_deep}, Embedding Density \cite{fishyscapes}, and Softmax Entropy \cite{MSP}. If a method has multiple variants, we show only the best. 

We also compare our method against Image Re-synthesis \cite{epfl}, as our framework builds upon it. Note that an official submission for this approach has not been done for Fishyscapes. As such, we implemented our own version of Image Re-synthesis and submitted it to the benchmark for comparison, which we named Image Resynthesis++.  
Appendix \ref{app:image_synthesis} contains details about our implementation and a quantitative comparison against the original work.

In order to ensure a fair comparison between the aforementioned methodologies, we divide the baselines into two groups depending on whether they require retraining the segmentation network or not. This split is intended to differentiate between methods that compromise segmentation accuracy to detect anomalies and methods that work as a wrapper to state-of-the-art (SOTA) segmentation models. To measure the difference between these two groups, we also report each method's mean intersection over union (mIOU) for all Cityscapes classes. This ensures that each approach produces competitive semantic segmentation predictions, while still detecting the anomalous instances.  

\subsection{Results}
\label{exp:results}
Table \ref{tab:results} shows quantitative comparisons between our proposed framework and the baselines discussed in Sec. \ref{exp:set_up} for the FS benchmark test sets. 

We first compare our approach against existing methodologies that do not jeopardize the performance of the segmentation model. 
Within this sub-group, our technique outperforms all previous best methods for the three datasets. Specifically, our method significantly improved the AP on all datasets. Additionally, the approach reduce the FPR95 for FS L\&F and FS Web by 50\%, while having comparable performance on this metric against the previous best method in FS Static. These results show the value of our contributions when grouped together. A detailed ablation study that quantifies the improvements of each added component can be found under Sec \ref{dis:ablation}.  

We then compare our proposed approach against methods that impact the segmentation network performance. In this comparison, our model had the best performance on FS L\&F in both AP and FPR95. Additionally, our approach achieves the second-best AP, behind Outlier Head \cite{anomaly_score} by 16\% and 4\% in FS Static and FS Web respectively. 
Our framework achieves comparable state-of-the-art performance on anomaly segmentation, while still maintaining state-of-the-art performance in semantic segmentation.  
Our pipeline accomplishes these results by working as a general framework which could potentially be implemented on top of different segmentation models (Sec. \ref{dis:generalization} and shows the framework's ability to generalize to different segmentation and synthesis networks). 
Note that one of the limitations of wrapper methods is their extended running time in comparison to methods that predict anomaly and segmentation in a single model. Details about inference time for our framework can be found in Appendix \ref{app:time}.

It is important to note that our technique is the only method that achieves top-2 performance in the Fishyscapes benchmark, showing the generalization ability of our method in different test sets. 
In comparison, previous methods, such as Dirichlet DeepLab \cite{prior_entropy}, has high AP in FS L\&F, but low AP in FS Static and FS Web. 
Similarly, Outlier Head \cite{anomaly_score} has a high AP in FS Static and FS Web, but the method does not generalize well for FS L\&F.

Qualitative comparison between our proposed framework and baselines for uncertainty methods \cite{MSP} and image-resynthesis methods \cite{epfl} can be found under Appendix \ref{app:examples}.

\section{Discussion}
\label{discussion}
When comparing all methods in Table~\ref{tab:results}, excluding ours, one could come to the conclusion that there is a trade-off between segmentation performance and anomaly detection. Our framework invalidates this hypothesis and shows that state-of-the-art anomaly detection does not have to come at a cost of segmentation quality. In the following, we discuss key insights that contribute to this result.

\subsection{Ablation Study}
\label{dis:ablation}
Table \ref{tab:ablation} provides results for an ablation study analyzing the contribution of individual components in the proposed method. 
We first find that both our training data generator and adding the uncertainty maps have significant improvements to the framework. The improvement due to the training data confirms the importance of covering all three anomaly scenarios of Figure~\ref{fig:anomaly_scenarios}, as opposed to just the one in \cite{epfl}. The further improvement due to uncertainty maps confirms our hypothesis that resynthesis and uncertainty carry complementary information, and combining these approaches leads to overall better performance. 
Note that removing the training data generator entails that the dissimilarity network does not train to use the uncertainty maps. 

We also find that combining the uncertainty maps (softmax entropy, softmax distance, and perceptual difference) with the output of the dissimilarity model greatly reduces the FPR95 for both datasets. This indicates that our dissimilarity module, like many deep CNNs, tends to be overconfident in its predictions~\cite{Calibration}. The ensemble helps to reduce this issue.
The FPR95 improvement does not correlate to AP as we see a drop in FS Lost \& Found. This drop is expected as we are combining our framework's prediction with a weaker detector in order to reduce FPR95 for safety-critical applications. Nonetheless, we see a boost in AP for FS Static. We explain this increment as FS Static artificially blends anomaly objects into urban landscapes images. As such, uncertainty methods outperform in these images as it is easier to detect the overlay of the objects.
In general, we expect a small drop in performance by AP using ensemble predictions, but a much larger improvements for FPR95. 

Finally, we observe that the ensemble has more consistent performance across trainings on all metrics.
This is expected as the impact of the dissimilarity module training on the overall performance of the ensemble is more limited than in the standalone case. Nonetheless, it is also a desirable property when deploying in safety critical applications.

\begin{table}[]\centering
\setlength{\tabcolsep}{3.5pt}
\begin{tabular}{lcccc}\toprule
\multirow{2}{*}{\textbf{Method}} &\multicolumn{2}{c}{\textbf{FS L\&F}} &\multicolumn{2}{c}{\textbf{FS Static}} \\
&$\uparrow$AP &$\downarrow$FPR95 &$\uparrow$AP &$\downarrow$FPR95\\ \toprule
Full Framework &$55 \pm 5$ &$\bm{40 \pm 5}$ &$\bm{62 \pm 5}$ &$\bm{26 \pm 1}$\\
w/o ensemble &$\bm{58 \pm 9}$ &$66 \pm 8 $&$57 \pm 6$ &$41 \pm 12$ \\
w/o unc. maps &$39 \pm 9$ &$64 \pm 10$ &$38 \pm 8$ &$51 \pm 4$ \\ 
w/o data generator &\multirow{2}{*}{$15 \pm 5$} &\multirow{2}{*}{$63 \pm 12$} &\multirow{2}{*}{$14 \pm 4$ }&\multirow{2}{*}{$57 \pm 11$} \\ \& w/o unc. maps \\
\midrule
Image Resyn.++ &$6 \pm 1$ &$48 \pm 12$ &$8 \pm  1$ &$63 \pm 18$\\
\bottomrule
\end{tabular}
\caption{\textbf{Ablation Study}. Ensemble, uncertainty maps, and data generators contribute to better overall performance. Image Resynthesis++ is used for comparison as our method builds upon it. Results are given as average and standard deviation over five random weight initializations.}
\label{tab:ablation}
\vspace{-5mm}
\end{table}

\subsection{Framework Generalization}
\label{dis:generalization}
The dissimilarity module in the proposed framework serves as a wrapper method for the segmentation and synthesis networks. In other words, the architecture shown in Figure \ref{fig:dissimilarity} is independent of the specific segmentation and synthesis approaches, as long as the segmentation network has a softmax layer as its output. 

We validate this generalization ability by re-training our dissimilarity network with different segmentation and synthesis techniques than the ones presented in Sec. \ref{exp:set_up}. Specifically, we chose ICNet as our segmentation module \cite{icnet} and SPADE as our synthesis module \cite{SPADE}. 
We selected these networks to create a lighter version of our pipeline, which we called \textit{Ours Light}. These lighter networks are significantly faster than the ones introduced in Sec. \ref{exp:set_up}, but with lower performance for their respective tasks. 

\begin{table}[]\centering
\setlength{\tabcolsep}{2.6pt}
\begin{tabular}{lcccccc}\toprule
\multirow{2}{*}{\textbf{Method}} &\multicolumn{2}{c}{\textbf{FS L\&F}} &\multicolumn{2}{c}{\textbf{FS Static}} &\textbf{CS} \\
&$\uparrow$AP &$\downarrow$FPR95 &$\uparrow$AP &$\downarrow$FPR95 &$\uparrow$mIOU \\\midrule
Im. Resyn.++ &5.7 &47.7 &8.0 &62.7 &\textbf{83.5} \\ 
Ours Light &36.0 &46.4 &33.4 &36.1 &70.6 \\
Ours &\textbf{55.1} &\textbf{39.6} &\textbf{61.5} &\textbf{25.6} &\textbf{83.5} \\
\bottomrule
\end{tabular}
\caption{\textbf{Performance comparison between best and lighter frameworks}. Our framework generalizes well to different segmentation and synthesis, even those with lower performance.}
\label{tab:light}
\vspace{-5mm}
\end{table}

Table \ref{tab:light} shows the performance of our best and lighter frameworks, as well as Image resynthesis++ as our baseline. Ours Light significantly outperforms the Image resynthesis++ baseline, even though this lighter version is using segmentation and synthesis modules with lower performance (i.e 83.5\% vs 70.6\% class mIOU on Cityscapes). 
This study demonstrates that not only the dissimilarity network generalizes to different segmentation and synthesis networks, but it also performs well even when the segmentation and synthesis networks produce lower quality outputs. 

Table \ref{tab:light} also indicates a direct correlation between the performance of the segmentation and synthesis modules and the anomaly detection accuracy. \textit{Ours} outperforms \textit{Ours Light} in all metrics, as \textit{Ours} uses state-of-the-art networks. 
These results agree with our intuition that re-synthesis methods are highly related to the quality of segmentation and synthesis networks. The better the predictions of these two modules, the easier the dissimilarity module differentiates between the input and synthesized image.  
As these networks improve in the upcoming years, we expect an improvement in performance for our anomaly detector framework.
An inference time analysis between Ours and Ours Light can be found in Appendix \ref{app:time}.

\section{Conclusion}
We investigate pixel-level anomaly detection for complex driving scenes (i.e urban landscapes).
We design an anomaly segmentation framework that combines two complementary approaches to anomaly detection: uncertainty and re-synthesis methods.
Specifically, our framework leverages uncertainty measurement maps (i.e. softmax entropy, softmax distance and perceptual differences) to guide a dissimilarity network to find the differences between the input image and a generated image from the predicted semantic map.
The presented approach significantly outperforms both re-synthesis and uncertainty based methods on the Fishyscapes benchmark, where it is the best overall method across datasets. It does not put any constraint on the segmentation network, and therefore can be used with any already trained state-of-the-art segmentation model. In fact, we demonstrate that our method also works well with lighter segmentation and synthesis networks, making it ready for deployment in autonomous machines.

{\small
\bibliographystyle{ieee_fullname}
\bibliography{main}
}

\cleardoublepage
\appendix
\section{Appendix}
\label{appendix}
\subsection{Implementation of Dissimilarity Network}
\label{app:dissimilarity}
The dissimilarity module is derived from six unique components: two encoding architectures, one fusion module, and three decoder blocks. These components are re-used across the network to build the final architecture. Figure \ref{fig:dissimilarity} shows a high-level view of all the components interconnected.  

We follow the naming convention used in \cite{perceptual_diff} and Pix2PixHD \cite{Pix2PixHd} to explain each architecture. Let $ck-sn$ denote a 3x3 Convolution-RELU layer with $k$ filters and stride $n$. $dk$ denote a 7x7 Convolution-RELU layer with $k$ filters and stride 1. $m2$ denotes a 2x2 max pooling layer. $sp-19$ denotes a SPADE normalization-SELU layer \cite{SPADE,selu}, which uses the 19 channels from the predicted semantic map as one of its inputs. $tk$ denotes a 2x2 transposed convolution with $k$ filters. Lastly, $r2$ denote a 1x1 Convolution layer with 2 filters 

\textbf{Input and Generated Image Encoder}. The first encoder uses the same base architecture as VGG16 \cite{vgg}:  $c64-s1$, $c64-s1$, $m2$, $c128-s1$, $c128-s1$, $m2$, $c256-s1$, $c256-s1$, $c256-s1$, $m2$, $c512-s1$, $c512-s1$, $c512-s1$. This architecture outputs four feature maps, one after each resolution. In other words, we output the features after the max pooling layer, as well as the final feature map from the encoder. This encoder shares weights for encoding both the input and synthesized image. 

\textbf{Semantic and Dispersion Maps Encoder}. This other encoder architecture is divided as: $d32$, $c64-s2$, $c128-s2$, $c256-s2$. The encoder outputs the feature maps after each convolutional layer block. Note that we use different weights to encode semantic information and uncertainty information. 

 \textbf{Fusion Module}. The fusion module concatenates the input, synthesized, and semantic feature maps at each resolution level. We then run a 1x1 convolution to extract important areas in the map, as well as reducing complexity. Finally, we perform a point-wise correlation between the dispersion feature map and the resulting map from the 1x1 convolution. This module will output a total of four feature maps - one for each resolution.  
 
  \textbf{Decoder Blocks}. There are four decoder blocks used in the dissimilarity network. The first and second one follow the same structure: $c256-s1$, $sp-19$, $c256-s1$, $sp-19$, $t256$. The third one is divided as: $c384-s1$, $sp-19$, $c128-s1$, $sp-19$, $t1258$, while the last one follows: $c192-s1$, $sp-19$, $c64-s1$, $sp-19$, $r2$. The first decoder block takes the feature map from the lowest resolution. All subsequent decoder blocks take as input the concatenation of the feature map from the fusion module and the output of the previous decoder block.

The dissimilarity network was trained for fifty (50) epochs, using the Adam \cite{adam} solver and a learning rate of 0.0001. We reduce the learning rate on plateau with a patience of 10 epochs. Additionally, we augment the training images by flipping around the vertical axis and normalizing them using mean and standard deviation values from ImageNet \cite{imagenet}.

\subsection{Re-Implementation of Image Re-Synthesis}
\label{app:image_synthesis}
Image Re-synthesis \cite{epfl} is a synthesis-based framework for anomaly segmentation. It consist of a segmentation model \emph{S}, a synthesis model \emph{G}, and a discrepancy network \emph{D}. Given a natural image, the framework will first predict a semantic label map with \emph{S}. Then, the synthesis model \emph{G} will re-synthesis the semantic map and finally the discrepancy network \emph{D} detects meaningful distances caused by mislabeled objects by comparing the natural and synthesized images.
The method adopts Bayesian SegNet \cite{BayesNet} and PSP Net \cite{PSPNet} as its segmentation models and Pix2PixHD \cite{Pix2PixHd} as its synthesis model. 

Performance of re-synthesis methods in anomaly segmentation, such as Image Re-synthesis, are highly related to the quality of the segmentation and synthesis predictions. If we improve these  predictions, the discrepancy network will have an easier task detecting anomalies.

To ensure that performance differences between our method and~\cite{epfl} do not come from the differences in the segmentation or synthesis modules, we re-implemented Image Re-synthesis with state-of-the-art segmentation and synthesis networks. Specifically, we replace their segmentation and synthesis networks with the same networks used in our framework (\cite{DeepLabV3+Label} for segmentation and \cite{ccfpse} for synthesis). By doing so, the only differences between both methods are our contributions explained in Section \ref{methodolgy}. 

Table \ref{tab:baseline_comparison} shows the differences between our implementation (Image Resynthesis++) against the results presented in \cite{epfl}. In our experiments, we use the same datasets and metrics used in the original publication. Specifically, we use \textit{Lost \& Found} (L\&F) \cite{LostFound} (i.e. images in a driving environment with small road hazards) and \textit{Road Anomaly} \cite{epfl} (i.e online images with anomalous objects located on or near the road) as our datasets, as well as the area under the curve for the receiver operating curve (AUC ROC) as our performance metric. 
We also added a variation of \textit{Lost \& Found}, where we restrict evaluation to the road, as defined by the ground-truth annotations. 

Our implementation achieves better performance across the different datasets, thus concluding our Image Resynthesis++ to be a stronger baseline when compared to our framework. The final implementation was submitted to the Fishyscapes Benchmark (private test set) to ensure equal comparison in more challenging anomaly dataset, such as FS Lost \& Found and FS Static.  

\begin{table}[!htp]\centering
\setlength{\tabcolsep}{3.5pt}
\begin{tabular}{lcccc}\toprule
\multirow{3}{*}{\textbf{Method}} &\multirow{2}{*}{\textbf{L\&F}} &{\textbf{L\&F}} &\textbf{Road} \\
& &\textbf{(Road Only)} &\textbf{Anomaly}\\ 
&\multicolumn{3}{l}{\qquad \hspace{3pt} $\uparrow$ AUC ROC} \\ \midrule
Image Resynthesis &0.82 &0.93 &0.83 \\
Image Resynthesis++ &\textbf{0.93} &\textbf{0.99} &\textbf{0.86} \\
\bottomrule
\end{tabular}
\caption{\textbf{Image Resynthesis implementation comparison}. Image Resynthesis++ outperforms the results presented in \cite{epfl} using the same datasets (Lost \& Found, Road Anomaly) with the area under the curve for the receiver operating curve (AUC ROC) as the performance metric.}
\label{tab:baseline_comparison}
\end{table}

We use the AUC ROC as our performance metric in these experiments since it was the only metric reported in the original work. However, as mentioned in Section \ref{exp:set_up}, ROC metrics are not well-suited for highly imbalance problems, such as anomaly detection, as explained in \cite{roc}. Thus, for our main experiments, we use more reliable metrics for imbalance problems, such as average precision (AP) and false positive rate at 95\% true positive rate (FPR95). 

Note that the Road Anomaly Dataset was not used in our main experiments, as it only contains sixty (60) images, which are not enough to ensure proper generalization abilities within anomaly segmentation. Additionally, the annotations are not consistent for the anomaly objects. For example, a rock in the middle of the road is labeled as an anomaly. However, the same style of rock next to the road is classfied as an inlier.

\subsection{Computational Complexity}
\label{app:time}
Table \ref{tab:inference_time} shows the inference time for each module in the proposed approach. Note that the perceptual difference dispersion map also requires running a CNN. As such, we also added its inference time to the running complexity. 
The estimated times are the average of running an image one hundred (100) times in our framework. We use an NVIDIA 1080Ti GPU with 11GB GPU memory with an input resolution for each module as described in Sec. \ref{exp:set_up}. Additionally, we also evaluate the inference speed of our lighter framework (explained in Sec. \ref{dis:generalization}) as a comparison.

\begin{table}[!htp]\centering
\setlength{\tabcolsep}{3.5pt}
\begin{tabular}{lcccc}\toprule
\textbf{Module} &\textbf{Ours} &\textbf{Ours Light} &\textbf{Resolution} \\\midrule
Segmentation &1256 &47 &2048x1024 \\
Synthesis &192 &62 &1024x512 \\
Perceptual Difference &13 &13 &512x256 \\
Dissimilarity &53 &53 &512x256 \\ \midrule
\textit{Total (ms)} &\textit{1514} &\textit{175} &\textit{--}\\
\bottomrule
\end{tabular}
\caption{\textbf{Computational Complexity Study}. Inference time and input resolution for each module in the proposed framework. Average results for one hundred (100) experiments using NVIDIA 1080Ti GPU.}
\label{tab:inference_time}
\end{table}

\subsection{Ensemble with End-to-End Training}
\label{app:end-to-end}
As stated in Sec. \ref{met:ensemble}, there are benefits to be gained by combining the calculated uncertainty maps (i.e softmax entropy, softmax distance and perceptual difference) with the output of the dissimilarity network. 
In our main work, we combine these predictions using a weighted average, where the weights are selected empirically through a grid search. 
An alternative approach would be to learn these weights during the training of the dissimilarity network.
This type of training would entail adding a learnable parameter (scalar) for each prediction map at the end of the dissimilarity network. Then, the model can optimize the weights to produce an  end-to-end ensemble prediction.

Table \ref{tab:end-to-end} compares the results between empirical and lernable weights using the validation set for FS Lost \& Found and FS Static.
We first find a discrepancy in AP performance. The end-to-end ensemble performs better in FS L\&F while empirical weights perform slightly better in FS Static. 
As stated in Sec. \ref{dis:ablation}, we explain that the ensemble prediction with empirical weights outperforms in FS Static since it is easier for uncertainty methods to detect artificially blended objects. This behavior is less evident in the end-to-end training since the network optimizes the weights before generating a final prediction. 
In general, we expected end-to-end ensemble to outperform emperical weights in AP as the network optimizes its weights more efficiently.  

The biggest insight from this comparison is shown when comparing the FPR95. In this metric, The end-to-end training significantly decreases performance when compared to empirical weights. 
These results are consistent with our ablation study in Sec. \ref{dis:ablation}, where we show that deep CNNs (e.g. dissimilarity module) tend to be overconfident with its prediction. 
Thus, by training in an end-to-end fashion, we are still prone to generating overconfidence outputs. 
As we intent to deploy our framework in safety critical environments (e.g. autonomous driving), we selected the empirical weights as our best model.

\begin{table}[]\centering
\setlength{\tabcolsep}{3.5pt}
\begin{tabular}{lcccc}\toprule
\multirow{2}{*}{\textbf{Method}} &\multicolumn{2}{c}{\textbf{FS L\&F}} &\multicolumn{2}{c}{\textbf{FS Static}} \\
&$\uparrow$AP &$\downarrow$FPR95 &$\uparrow$AP &$\downarrow$FPR95\\ \toprule
Ours w grid search &55.1 &\textbf{39.6} &\textbf{61.5} &\textbf{25.6} \\
Ours w end-to-end &\textbf{59.6} &58.6 &61.1 &37.3 \\
\bottomrule
\end{tabular}
\caption{\textbf{Ensemble Prediction Comparisons}. Grouping the predictions through end-to-end training shows comparable results in AP against empirically selected weights (grid search). However, end-to-end training increases significantly the FPR95 as the network gets overconfident with its predictions.}
\label{tab:end-to-end}
\end{table}


\subsection{Example Predictions}

\label{app:examples}
\begin{figure*}[!th]
\begin{center}
   \includegraphics[width=1.0\linewidth]{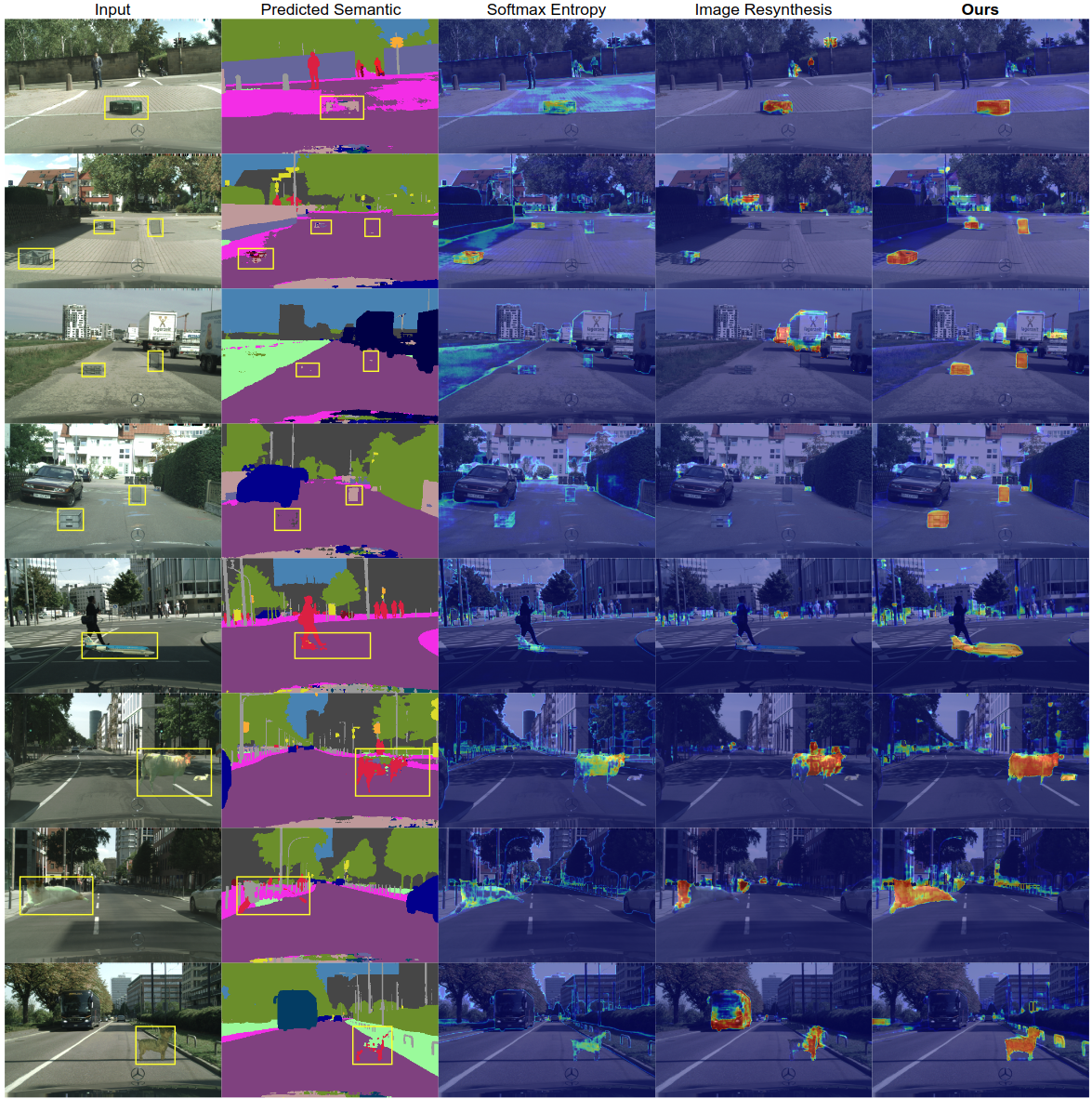}
\end{center}
   \caption{\textbf{Framework example predictions}. Qualitative comparison between proposed framework and baseline for uncertainty methods \cite{MSP} and image-resynthesis methods \cite{epfl}. The proposed framework outperforms both previous methods detecting anomalies instances. The first five images are from the FS Lost \& Found, while the next five are from FS Static. Pixels labeled as \textit{void} are excluded from the prediction visualizations for the three methods shown, as they are also excluded in the anomaly benchmarks.}
\label{fig:examples}
\end{figure*}

\begin{figure*}[!th]
\begin{center}
   \includegraphics[width=1.0\linewidth]{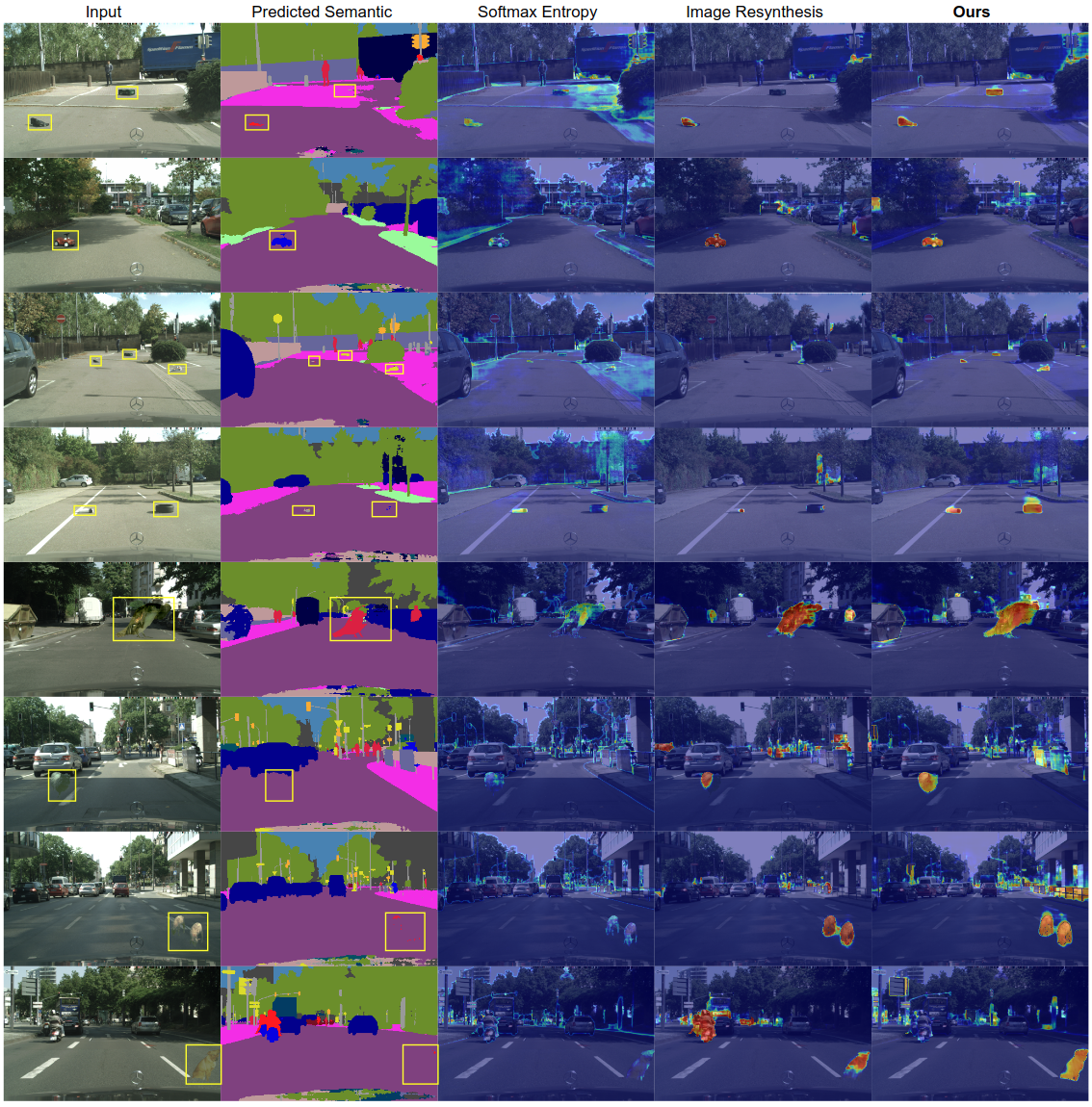}
\end{center}
   \caption{\textbf{Additional predictions examples.} Our framework reliable detects all three outcomes when a segmentation network encounters an anomalous instances, as explained in Figure \ref{fig:anomaly_scenarios}. First four images are from FS Lost \& Found, while the next four are from FS Static. Softmax Entropy \cite{MSP} and Image Resynthesis \cite{epfl} are also shown as reference. Pixels labeled as \textit{void} are excluded from the prediction visualizations for the three methods shown, as they are also excluded in the anomaly benchmarks.}
\label{fig:add_examples}
\end{figure*}

Figure \ref{fig:examples} and Figure \ref{fig:add_examples} display example predictions of our approach in validation images from FS Lost \& Found and FS Static. Additionally, we show a qualitative comparison between our technique, an uncertainty estimation method (Softmax Entropy \cite{MSP}), and an image re-synthesis method (Image Re-synthesis \cite{epfl}). These images emphasizes the robustness of our framework for all anomalies scenarios (Figure \ref{fig:anomaly_scenarios}), in comparison with previous methods. Note that the anomaly detection framework did not train with any of the anomalous instances, and they are seen for the first time during testing. 

\begin{figure*}[!th]
\begin{center}
   \includegraphics[width=1.0\linewidth]{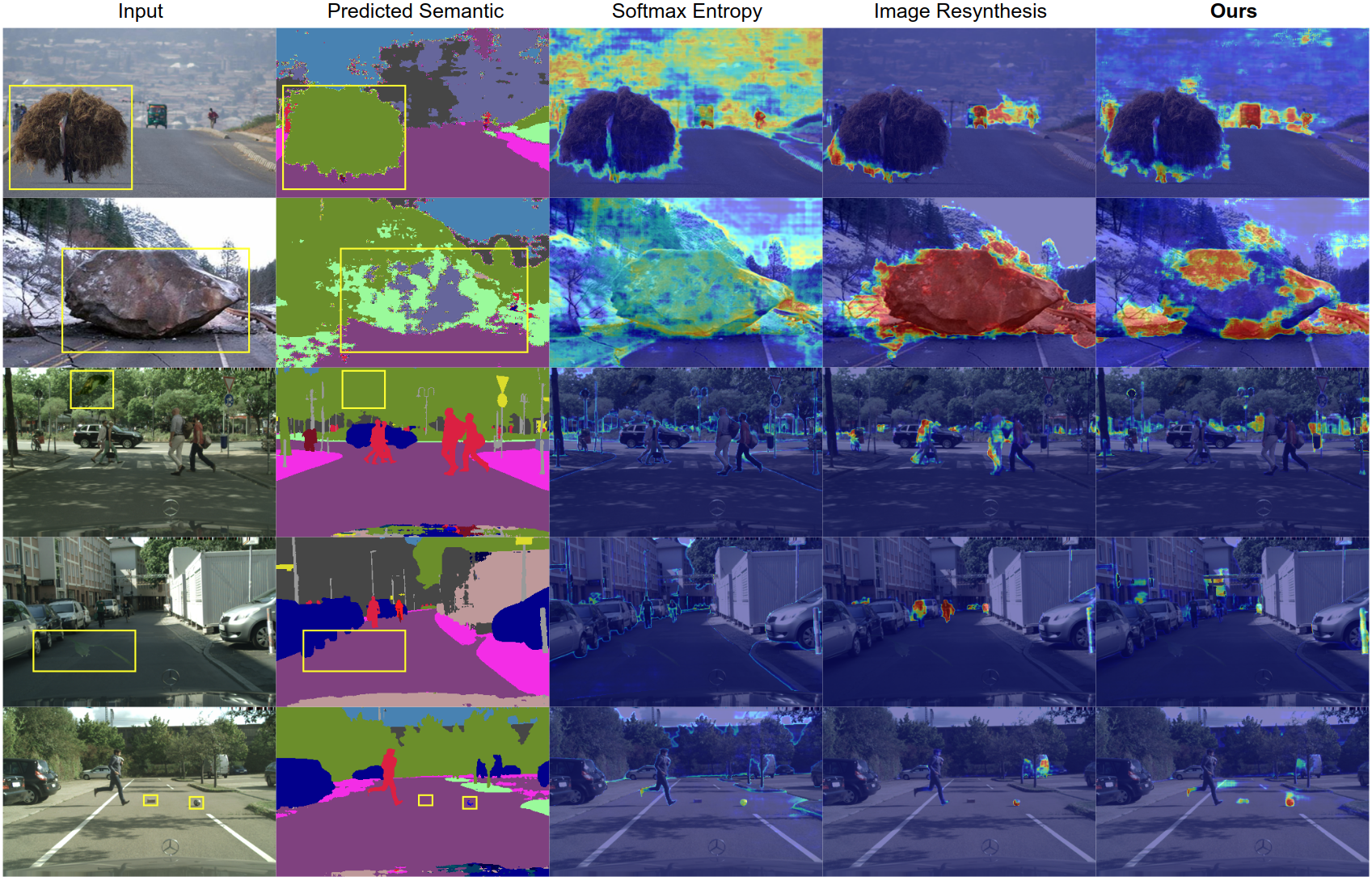}
\end{center}
   \caption{\textbf{Failure cases.} Our framework still fails at detecting some challenging anomaly instances. Common errors are derived scenes that differ urban landscape (top two images), anomaly instances that blend well with the background (middle two images) or small anomalous objects that only cover few pixels in the image. Softmax Entropy \cite{MSP} and Image Resynthesis \cite{epfl} are also shown as reference.}
\label{fig:failure}
\end{figure*}

Additionally, some failure cases are presented in Figure \ref{fig:failure}. Common errors are derived scenes that differ urban landscape, anomaly instances that blend well with the background or small anomalous objects that only cover few pixels in the image.

\newpage

\newpage

\newpage

\end{document}